\documentclass[journal]{IEEEtran}

\usepackage{algorithmic}
\usepackage{algorithm} 
\usepackage{booktabs}
\usepackage{graphicx}
\usepackage{multirow} 

\usepackage{tabularx} 
\usepackage{ragged2e} 
\usepackage{booktabs}

\usepackage{color}
\usepackage{hyperref}
\hypersetup{hidelinks,
	colorlinks=true,
	allcolors=black,
	pdfstartview=Fit,
	breaklinks=true}

\usepackage{amsmath}

\usepackage{graphicx}

\begin{document}
%
\title{AAPMT: AGI Assessment Through Prompt and Metric Transformer}
%
%
%

\author{Benhao Huang, Shanghai Jiao Tong University
        ~\IEEEmembership{}
\thanks{Benhao Huang is an undergraduate student from Shanghai Jiao Tong University. E-mail:hbh001098hbh@sjtu.edu.cn}
}

\maketitle

\begin{abstract}
The emergence of text-to-image models marks a significant milestone in the evolution of AI-generated images (AGIs), expanding their use in diverse domains like design, entertainment, and more. Despite these breakthroughs, the quality of AGIs often remains suboptimal, highlighting the need for effective evaluation methods. These methods are crucial for assessing the quality of images relative to their textual descriptions, and they must accurately mirror human perception.
Substantial progress has been achieved in this domain, with innovative techniques such as BLIP and DBCNN contributing significantly. However, recent studies, including AGIQA-3K, reveal a notable discrepancy between current methods and state-of-the-art (SOTA) standards. This gap emphasizes the necessity for a more sophisticated and precise evaluation metric.
In response, our objective is to develop a model that could give ratings for metrics, which focuses on parameters like perceptual quality, authenticity, and the correspondence between text and image, that more closely aligns with human perception. 
In our paper, we introduce a range of effective methods, including prompt designs and the Metric Transformer. The Metric Transformer is a novel structure inspired by the complex interrelationships among various AGI quality metrics. The code is available at \href{https://github.com/huskydoge/CS3324-Digital-Image-Processing/tree/main/Assignment1}{here}.
\end{abstract}


%
\IEEEpeerreviewmaketitle

\section{Implementation of Model}
%
%
%
%
\subsection{Back Bone Model}
Blip is an excellent model for Text-Image Matching task. We utilized a recent work Image Reward \cite{xu2023imagereward}. Based on this work, we try to fine-tune its pretrained model on AGIQA-3K dataset and AIGCIQA2023 dataset correspondingly to gain results.

\subsection{Data Preparation}
Regarding data preparation, we partitioned the entire dataset into two distinct subsets: 80\% allocated for training and 20\% for testing. Crucially, we adopted a strategy where images generated from the same prompt are grouped together. This approach adheres to the "Content Isolation" principle, ensuring that the results derived from the test set are more reliable. 
However, due to time constraints, we maintained a consistent random seed for the dataset split and did not conduct cross-validation experiments. We acknowledge this as a limitation and plan to incorporate cross-validation in future iterations of our research to further validate our findings. But a simple robust test experiment will be included in Appendix. \label{weak1}

\subsection{Assess Text-Image Correspondence}
\begin{figure}[!htbp]
    \centering
    \includegraphics[scale=0.41]{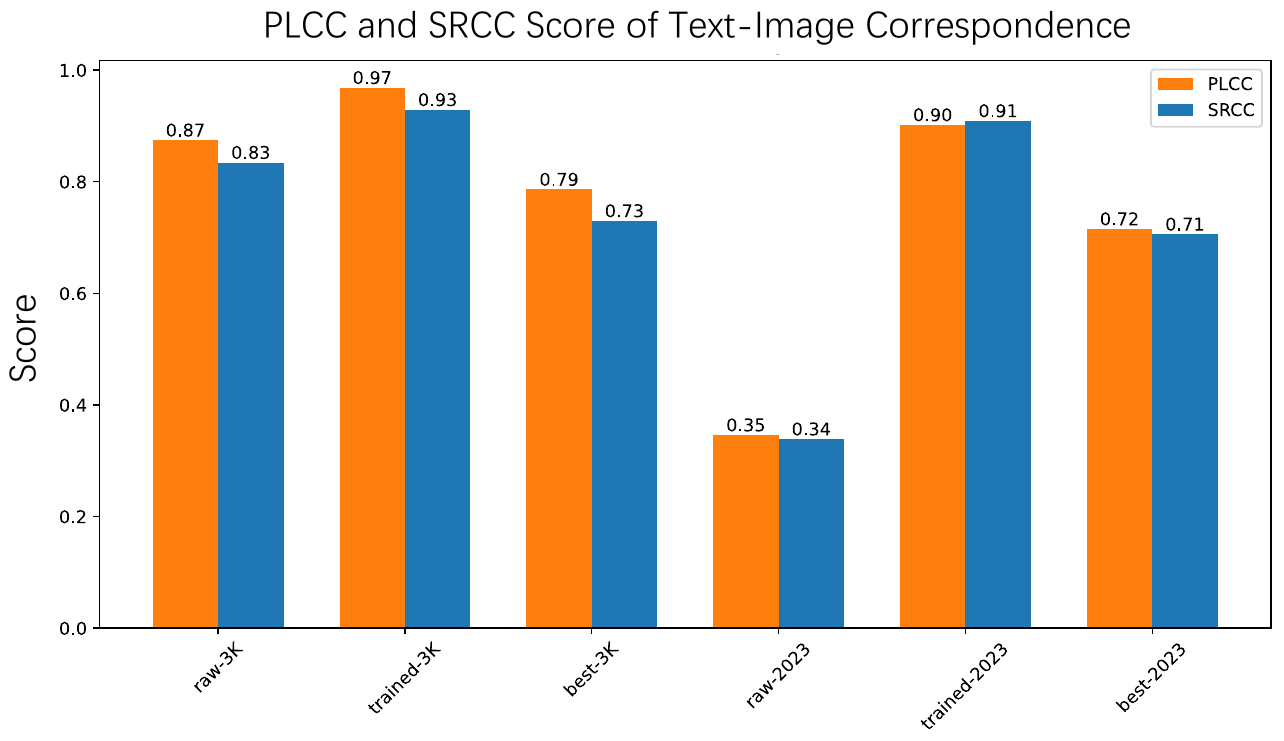}
    \centering
    \vspace{-8pt}
    \caption{Assess Text-Image Correspondence. '3K' signifies that the task is evaluated on the AGIQA-3K dataset, while '2023' indicates its evaluation on the AIGCIQA2023 dataset. The term 'raw' refers to using the Image Reward model that has not been trained on the specific dataset, in contrast, 'trained' denotes the use of models that have undergone training on these datasets. Additionally, we also present the best scores as reported in references \cite{3K} and \cite{2023}, corresponding to their respective datasets.}
    \label{TIC}
\end{figure}

Evaluating text-image correspondence can be conceptualized as a "Text-Image Match" problem: given an image and a paragraph of text as inputs, the model should output a score that quantifies the degree of match between the image and the text. This is precisely the model structure required for our task. Consequently, we employed the model provided by Image Reward\footnote{\href{https://github.com/THUDM/ImageReward}{https://github.com/THUDM/ImageReward}}, training it on our own dataset. In terms of hyperparameter selection, we largely adhered to the settings outlined in Image Reward\cite{xu2023imagereward}. Key hyperparameters are detailed in Table \ref{hyper}.
The reason for choosing for a higher number of epochs when training on the AGCIQA2023 dataset stems from the original model's performance on the AGIQA-3K dataset, where it was already competitive with numerous other Image Quality Assessment (IQA) models (SRCC =0.874 , PLCC = 0.834). However, a significant performance decline was observed when the model was tested on AIGCIQA2023, suggesting different properties between the two datasets. After conducting 10 epochs of training on AIGCIQA2023, we observed a significant improvement in the model's performance, but still suboptimal. Consequently, we increased the number of training epochs to 50, which we found sufficient to achieve satisfactory performance, as demonstrated in the figure \ref{TIC} .

\begin{table*}
  \caption{Hyperparameters Setting}
  \label{hyper}
  \centering
  \begin{tabular}{cccccccc}
  \toprule
  \textbf{Model} & \textbf{dataset} & \textbf{Training Epochs} & \textbf{Batch Size} & \textbf{Learning Rate} & \textbf{Adam Beta 1} & \textbf{Adam Beta 2} & \textbf{Adam Eps} \\
  \midrule
  ImageReward-v1.0 & AGIQA-3K & 10 & 32 & 5e-06 & 0.9 & 0.999 & 1e-8 \\
  ImageReward-v1.0 & AGCIQA2023 & 50 & 32 & 5e-06 & 0.9 & 0.999 & 1e-8 \\
  \bottomrule
  \end{tabular}
\end{table*}

\subsection{Assess Image Quality}

\subsubsection{A Naive Trial}
\begin{figure}[!htbp]
    \centering
    \includegraphics[scale=0.41]{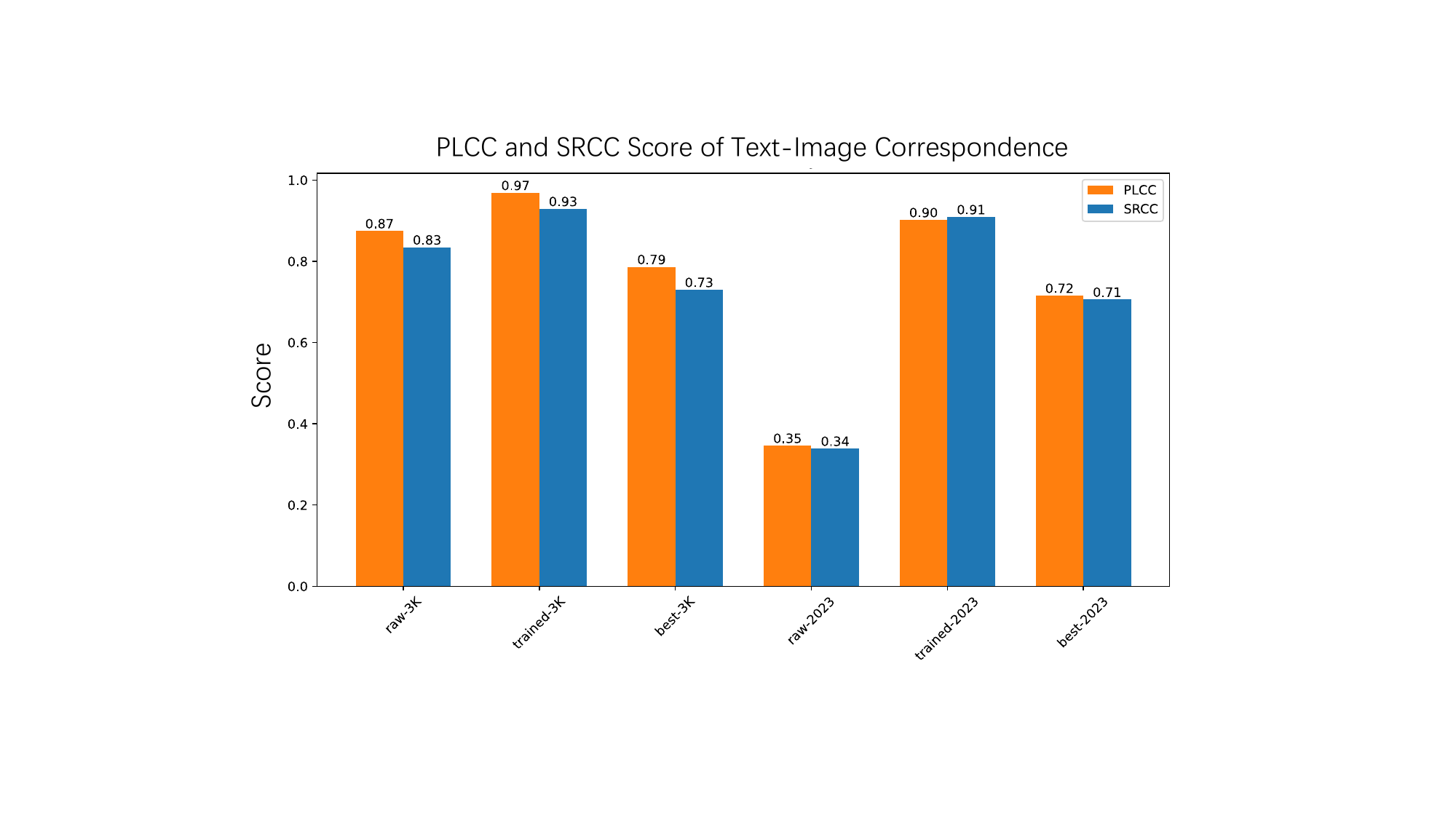}
    \centering
    \vspace{-15pt}
    \caption{Assess Image Quality by designing prompt. Same notations with figure \ref{TIC}.}
    \label{quality}
\end{figure}

Utilizing a model adept in the text-image matching task offers a straightforward yet effective strategy: modify the input prompt to explicitly state \textbf{"extremely high quality image, with vivid details"} and train the model on this dataset. The underlying principle is simple: the higher the output score from the model, the greater the correspondence between the image and the concept of the prompt "extremely high quality image, with vivid details." This implies that the image in question is of high quality.
Remarkably, this seemingly naive method yields surprisingly effective results. The performance of the model, as measured by Pearson Linear Correlation Coefficient (PLCC) and Spearman Rank Correlation Coefficient (SRCC), is excellent. The results' values  are detailed in the figure \ref{quality}.

\subsubsection{Change in Prompt Content}
Despite the initial success, we questioned whether this single prompt was optimal. To investigate further, we conducted additional experiments with three distinct prompts intended for image quality assessment. These prompts included: Prompt 1: \textit{"high quality image"}, Prompt 2: \textit{"extremely high quality image, with vivid details"}, and Prompt 3: \textit{"extremely high quality image, with high resolution"}. We reevaluated the model using these new prompts. The results, detailed in the table below, revealed some differences based on the specific prompt used. Notably, replacing "vivid details" with "high resolution" in Prompt 2 resulted in a decrease in both PLCC and SRCC scores. This finding suggests that the AI model may place greater emphasis on "vivid details" than "high resolution" when assessing image quality.

\begin{table}[!h]
  \caption{PLCC and SRCC Score of Image Quality with Different Prompts}
  \label{diff_prp}
  \centering
  \setlength{\tabcolsep}{5mm}{
  \begin{tabular}{cccc}
  \toprule
   & \textbf{Prompt1} & \textbf{Prompt2} & \textbf{Prompt3}  \\
  \midrule
  PLCC & 0.8812 & \textcolor{red}{0.8855} & 0.8594 \\
  SRCC & 0.8941 & \textcolor{red}{0.8976} & 0.8780 \\
  \bottomrule
  \end{tabular}}
\end{table}

Furthermore, the comparison between Prompt 1 and Prompt 2 shows a negligible difference, indicating that the phrase "high quality image" is the critical component in these prompts. Due to time constraints, we couldn't explore further possibilities, which would be an important avenue for future research.

\subsection{Assess Image Authenticity}
By following the same approach as used for assessing image quality, we selected the prompt \textbf{"very authentic image"} to evaluate the authenticity of images. Given that authenticity is not a metric included in the AGIQA-3K dataset, our evaluation was exclusively conducted on the AGCIQA2023 dataset. The results of this assessment are detailed in table \ref{authentic}.

\begin{table}[!h]
  \caption{PLCC and SRCC Score of Image Authenticity}
  \label{authentic}
  \centering
  \setlength{\tabcolsep}{4mm}{
  \begin{tabular}{cccc}
  \toprule
   & \textbf{raw-2023} & \textbf{trained-2023} & \textbf{best-2023}\cite{2023}  \\
  \midrule
  PLCC & 0.1286 & \textcolor{red}{0.8985} & 0.6807 \\
  SRCC & 0.1345 & \textcolor{red}{0.9082}  & 0.6701 \\
  \bottomrule
  \end{tabular}}
\end{table}

\section{Assessing Multiple Metrics with a Single Model}

So far, we have tried to use Image Reward model \cite{xu2023imagereward} to assess three metrics of AGI, which has a comparably excellent performance. Nevertheless, if we stick to this method, then we need to correspondingly train three models, namely models for quality, for authenticity as well as for text-image correspondence, which is quite inefficient and space-consuming.
In this section, we want to come up with more novel ideas to further enhance the efficiency of AGI assessment. Since AGIQA-3K only has two metrics, we choose AGCIQA2023 dataset to carry out our further experiments. The following subsections elaborate on some different approaches we have attempted and the thought process we underwent in trying to solve this problem.

\subsection{Second Training of the Model Initially Trained for a Different Task}
\label{sectrain}
A naive thought is that we could utilize the ability of a model which has been already trained for quality assessment to retrain it for another assessment task.
\begin{algorithm}[!htbp]
\caption{Second Training}
\begin{algorithmic}[1]
\STATE \textbf{Preparing the Dataset}: Dividing the Dataset into Two Parts, with 20\% for Testing and 80\% for Training
\STATE Model $\leftarrow$ Initialize from Image Reward Model Parameters
\STATE Train the Model for quality metrics
\STATE Save the checkpoint $ckpt_{quality}$
\STATE  Reinitialize a Model and load the checkpoint $ckpt_{quality}$
\STATE Train the Model for text-image correspondence metric

\end{algorithmic}
\end{algorithm}

It's obvious that this method didn't save anytime, since the model in this method still needs to be trained multiple times for different metrics. However, we could gain some insightful points from its results. Here, we focus solely on retraining the model for text-image correspondence that was previously trained for quality assessment and the opposite case.

\begin{table*}[!htbp]
   \caption{Results of Second Training. Align to Quality means training model for quality from a model which has been trained for assessing text-image correspondence task. Trained for Quality means merely training the model for assessing image quality task.}
    \centering
    \renewcommand{\arraystretch}{1.5}
    \setlength{\tabcolsep}{4mm}{
    
    \begin{tabular}{c|cc|cc|cc|cc}
    \midrule
    \multirow{2}*{Task} & \multicolumn{2}{c|}{Align to Quality} & \multicolumn{2}{c|}{Quality to Align} & \multicolumn{2}{c|}{Trained for Quality} & \multicolumn{2}{c}{Trained for Align} \\
    \cline{2-9} 
		  & PLCC  & SRCC & PLCC  & SRCC & PLCC  & SRCC & PLCC  & SRCC \\
		\hline
            AGCIQA2023-Quality & \textcolor{red}{0.9173}& \textcolor{red}{0.9194} &	0.6199	&0.6142 &	0.8855 &	0.8976 &	0.5316 &	0.5177 \\
            AGCIQA2023-Align &	0.6641 & 0.6645 &	\textcolor{red}{0.9145} &	\textcolor{red}{0.9160} & 	0.4652 &	0.5512 &	0.9018 &	0.9090 \\
    \bottomrule

    \end{tabular}}
    \label{second}
\end{table*}

From table \ref{second}, we have following discoveries: \begin{itemize}
    \item[1.] When retraining the trained model on a different task, its performance drops a lot on the original task. However, it's interesting to note that its performance, while reduced, remains better than that of a model which was never trained on the original task. For example, for AGCIQA2023-Align task, Align to Quality model still outperforms Trained for Quality.
    \item[2.] Comparing to merely training on task A(quality or alignment), training on another task B(alignment or quality) then retraining the model on task A could further improve the quality. \label{discorvery2}
\end{itemize}

From these phenomena, we could speculate that the optimal parameters space for quality task and text-image correspondence has relationship demonstrated by the following Venn diagram.

\begin{figure}[!htbp]
    \centering
    \includegraphics[scale=0.3]{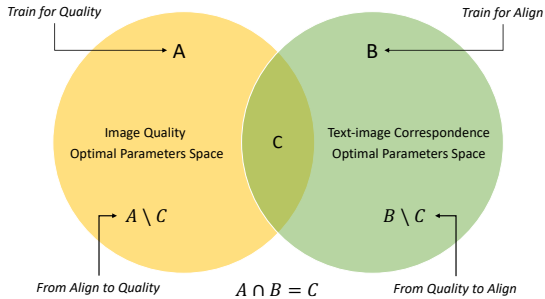}
    \caption{Venn diagram of Optimal Parameters Space. The notations here share the meaning with table \ref{second}.}
    \label{fig:enter-label}
\end{figure}

 When retraining the model, which has been trained to assess quality, for alignment task, then its parameters space will shift from $A$ to $B$, which will bring a drop of performance on assessing image quality. However, since there is an overlap between two optimal parameters' space, a second training model on new task $T$ could outperform the model trained only for task $T$, because in previous case, model already have a good knowledge encoded by the shared parameters space which is $A\cap B$. 
 Moreover, we could speculate that when training only for one task, it's hard for model to embody the shared parameter space, which makes sense intuitively. 
Therefore, when using same training strategy and hyperparameters, a second training model, which goes through a second training for task A and first training for task B, could perform  better on task A than those model merely trained for task A.

\subsection{One Step Even Further: Metric Transformer}

In subsection \ref{sectrain}, we explored various intriguing aspects of training models for different AGI assessment metrics. As previously noted, it appears that the optimal parameters for these metrics overlap to some extent. This observation leads to a plausible hypothesis: the metrics themselves might share semantic similarities. For example, an image with high text-image correspondence is likely to also be of higher quality, and vice versa. The one way we could utilize this phenomenon is to develop a model which could consider other metrics' influence when rating for a certain metrics. In order to make one metric assessment process have the information from other metrics, we develop \textbf{ Metric Transformer}, which utilizes the advantage of self attention\cite{vaswani2017attention} mechanism.

\begin{figure*}
    \centering
    \includegraphics[scale=0.6]{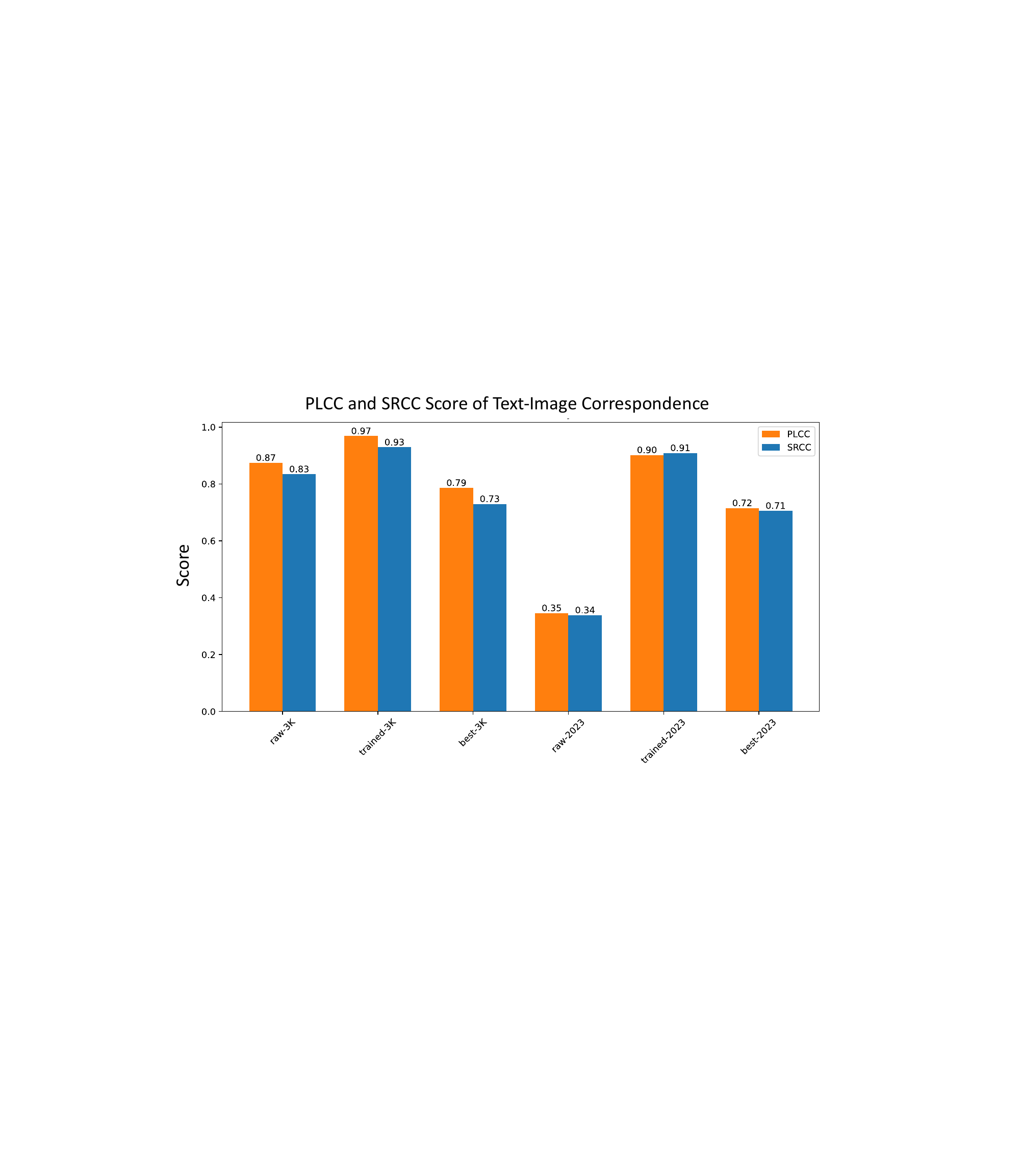}
    \caption{Diagram of the Metric Transformer. It's worth noting that the primary distinction between our model and Image Reward lies in the design of the final layer. We employ a model similar to the transformer (which typically has only one set of $W_{K,Q,V}$). This design allows the model to evaluate multiple image metrics concurrently, delivering impressive performance as demonstrated in Table \ref{mtres}.}
    \label{fig:enter-label}
\end{figure*}

Our model adopts the structure utilized in Image Reward\cite{xu2023imagereward} for text-image processing. We then replace the Multilayer Perceptron (MLP) layer of the original model with our proposed Metric Transformer. It's important to note that before feeding the text features into the Metric Transformer, we first pass them through a three-head transformer encoder. This initial step allows the model to learn the basic concepts of the three metrics from the training dataset.

These encoded features are then input into the Metric Transformer to calculate the scores for the three metrics correspondingly. If we denote these encoded features as $EF$, we then have:

\begin{equation}
    \begin{aligned}
Q_i &= W_{Q^i}^T \cdot EF \\
K_i &= W_{K^i}^T \cdot EF \\
V_i &= W_{V^i}^T \cdot EF
\end{aligned}
\end{equation}

Here $W_Q^i,W_K^i,W_V^i$ is the parameters of models which is used to get Query, Key and Value matrices.
Then we calculate the metric score as follows:

\begin{equation}
    S_i = \sum_{j=1}^{3} \operatorname{SoftMax}(\frac{Q_i K_j^T}{\sqrt{d_k}}V_j)
\end{equation}

Here $d_k$ is the dimension of $K_i$. Then we trained this model with exactly the same setting of Image Reward model trained on AGCIQA2023, and derived the results shown in table \ref{mtres} 

\begin{table*}[!h]
   \caption{Results of Assessing IQA Metrics Using Metric Transformer. Here the best score for a given metric on a task is designated by \textcolor{red}{RED}, while the second-best score is indicated by \textcolor{blue}{BLUE}. The results demonstrate that the overall performance of the Metric Transformer is definitively on par with that of Image Reward. It's important to emphasize that the score for Image Reward is actually derived from three models specifically trained for different tasks, whereas the Metric Transformer involves only a single model. This point should be particularly considered when evaluating their performance. }
    \centering
    \renewcommand{\arraystretch}{1.5}
    \setlength{\tabcolsep}{4mm}{
    
    \begin{tabular}{c|cc|cc|cc}
    \midrule
    \multirow{2}*{Task} & \multicolumn{2}{c|}{Best in AIGCIQA2023} & \multicolumn{2}{c|}{Trained Image Reward} & \multicolumn{2}{c}{Metric Transformer}\\
    \cline{2-7} 
		  & PLCC  & SRCC & PLCC  & SRCC & PLCC  & SRCC  \\
		\hline
            AGCIQA2023-Quality & 0.8402 & 0.7961 &	\textcolor{red}{0.8855}	& \textcolor{red}{0.8976} &	\textcolor{blue}{0.8819} &	\textcolor{blue}{0.8852} \\
            AGCIQA2023-Align &	0.7153 & 0.7058 &	\textcolor{blue}{0.9018} &	\textcolor{red}{0.9090} &  \textcolor{red}{0.9060} &	\textcolor{blue}{0.9068}  \\
            AGCIQA2023-Authenticity &	0.6807 & 0.6701 &	\textcolor{blue}{0.8985} & \textcolor{red}{0.9082} & 	\textcolor{red}{0.9112} &	\textcolor{blue}{0.9075}  \\
    \bottomrule

    \end{tabular}}
    \label{mtres}
\end{table*}

\begin{figure}[!h]
    \centering
    \includegraphics[scale=0.5]{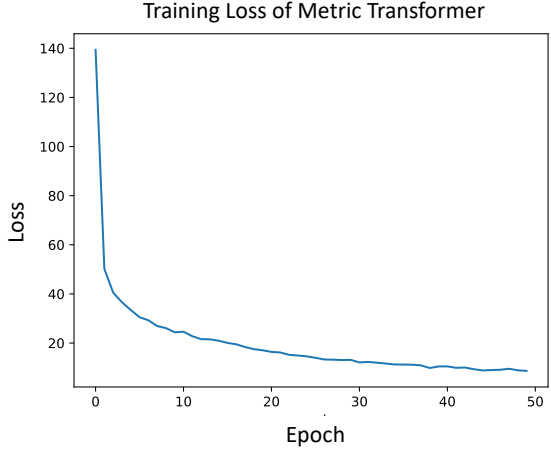}
    \caption{Training Loss of Metric Transformer in 50 Epochs}
    \label{MT_loss}
\end{figure}

We also depict the loss trajectory during the training of the Metric Transformer in Figure \ref{MT_loss}. From the plot, it's evident that the loss continues to show a downward trend even after 50 epochs. However, considering the constraints of computational resources and time, we made the decision to halt training at this juncture.

\section{Conclusion}
In this paper, we leverage the model from previous work, Image Reward \cite{xu2023imagereward}, to successfully evaluate multiple metrics relevant to AI-generated images. These metrics include perceptual quality, authenticity, and text-image correspondence. Beyond testing the naive method using pre-designed prompts, we further analyze the interplay between these metrics and hypothesize that they mutually influence each other. Drawing on these insights, we propose a novel model structure, the Metric Transformer, specifically designed to accelerate model training as well as improve model performance in AGI assessment. This model is capable of rating multiple metrics in a single run, displaying high correspondence with human evaluation scores. We delve deeper into our insights and render complementary experiments in the Appendix.


%

\appendices
\section{Test Model Performance with Different Seed}
As we mentioned in \ref{weak1}, we try to prove that the excellent performance of the model is not determined by some random factors. So we choose three seeds to initialize the dataset and model to ensure this point. We train and test on AGIQA-3K dataset.

\begin{table*}[!htbp]
   \caption{Results of Different Initialization. To prove the excellent performance of prompt design method is robust, we test it on three randomly picked different seed. We could see that there is no obvious performance gap between different seed, which is used for model parameters initialization and dataset shuffle.}
    \centering
    \renewcommand{\arraystretch}{1.5}
    \setlength{\tabcolsep}{5mm}{
    
    \begin{tabular}{c|cc|cc|cc}
    \midrule
    \multirow{2}*{Task} & \multicolumn{2}{c|}{Seed = 42} & \multicolumn{2}{c|}{Seed = 100} & \multicolumn{2}{c}{Seed = 200} \\
    \cline{2-7} 
		  & PLCC  & SRCC & PLCC  & SRCC & PLCC  & SRCC \\
		\hline
            AGIQA-3K Quality & 0.9694 & 0.9329 &	0.9530	& 0.9119 &	0.9577 &	0.9175 	 \\
            AGIQA-3K Align &	0.9695 & 0.9392 &	0.9631 &	0.9195 & 	0.9714 &	0.9363 	\\
    \bottomrule

    \end{tabular}}
    \label{second}
\end{table*}
\begin{figure*}[!h]
    \centering
    \includegraphics[scale=0.55]{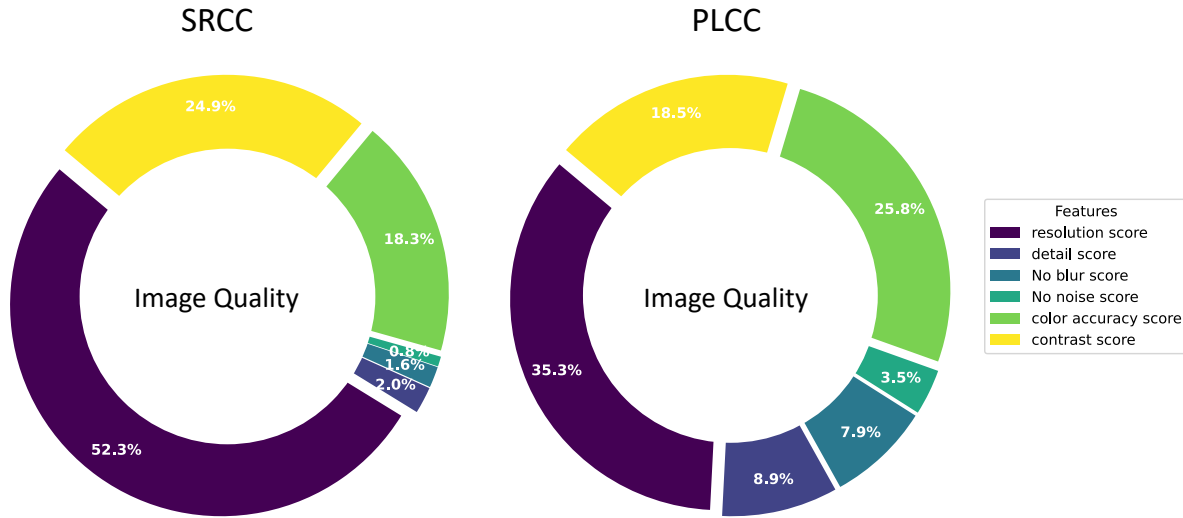}
    \caption{Pie Chart Showing the Corresponding Ratio of Different Child Metrics. As we could see from the chart, there truly exists some difference between different child metrics, which could be explored further.}
    \label{pie}
\end{figure*}

\section{More Discussion and Future Works}
\subsection{Design the Loss for Multiple Metrics}
Regarding training a model for multiple metric assessment, we have thought that it's possible for us to design a dynamic loss, which is :

\begin{equation}
    \mathcal{L} = \sum_{i=1}^n \alpha_i \cdot Loss_i, \quad 
    \text{where} \sum_{i=1}^n \alpha_i = 1
\end{equation}

Here $\alpha_i$ is called loss parameters, and $loss_i$ is the model's loss on metric $i$. During the training process, we update parameters $\alpha_i, i = 1,2,3,..., n$ as follows:

\begin{algorithm}[!htbp]
\caption{Updating Loss Parameters}
\begin{algorithmic}[1]
\STATE Randomly Initialize $\alpha_i$, which satisfies $\sum_i \alpha_i = 1$
\STATE previous $\alpha_i$ $\leftarrow \alpha_i $
\FOR{each step}
\FOR{ each metric $i$}
    \STATE $loss_i \leftarrow$ model(input, $\text{metric}_i$)
\ENDFOR
\STATE  $\alpha_i' \leftarrow$  $\frac{ \text{previous } \alpha_i \cdot loss_i}{\sum_{i=1}^n \alpha_i \cdot loss_i}$

\STATE previous $\alpha_i \leftarrow \alpha_i$
\STATE $\alpha_i \leftarrow \alpha_i'$
\ENDFOR
\end{algorithmic}
\end{algorithm}

Due to limited time, we haven't fully explore it yet. Thus we simply include it in Appendix here.

\subsection{Disentangle the sub-metric of Image Quality}
As discussed in Section \ref{sectrain}, there could be interactions between different metrics. This observation leads to an intriguing question: is it possible to disentangle a parent metric into several child metrics? For instance, consider image quality. Despite numerous exceptional studies \cite{karras2020analyzing}\cite{ding2020image} that concentrate on enhancing image quality, to the best of my knowledge, the disentanglement of image quality into subsidiary metrics remains largely unexplored. Consequently, we propose the following approach: We begin by identifying some crucial metrics pertinent to image quality.

Firstly, we enumerate several significant metrics that are pertinent to image quality. 

\begin{itemize}
\item \textbf{Resolution}: This is a basic measure of image clarity. A higher resolution entails richer details and more natural color gradients in the image.
\item \textbf{Vivid Details}: This refers to whether the elements in the image are clearly depicted without any blurring or distortion. This includes whether the edges of the image are sharp, and whether textures, shadows, and highlights are clearly visible.
\item \textbf{No Blur}: This refers to whether the image is clear, without any blurring or distortion. This could be due to the camera's focus being off, or blurring occurring in the image creation process.
\item \textbf{Color Accuracy}: This refers to whether the colors in the image truthfully reflect the original scene. Color accuracy can be influenced by various factors, including lighting conditions, camera settings, and image editing.
\item \textbf{Contrast}: This refers to the degree of difference between the bright and dark parts of an image. Good contrast can help to highlight the details and depth of the image.
\item \textbf{Noise Level}: This refers to random variations in color in the image, usually appearing as grainy or speckled discoloration. High-quality images should minimize noise as much as possible.
\end{itemize}

Next, we consider the question of quantification: How can we measure this pattern? The methodology we've developed employs specifically designed prompts. We utilize the score obtained from the prompt "extremely high quality image, with vivid details" as the base score, denoted as \( S_{base} \). The scores from other prompts - such as "high resolution," "with vivid details," "without blur," "without noise," "with accurate color," and "with high contrast" - are denoted as \( S_i \). We then calculate their relative significance in assessing image quality by determining their counting ratio:
\begin{equation}
    ratio_i = \frac{{1}/{(S_{base}-S_i)}}{\sum_{i=1}^n\frac{1}{S_{base}-S_i}}
\end{equation}

In the subsequent experiment, we utilized a model that was specifically trained to assess image quality using the prompt "high quality image." The results showed that \( S_{base} > S_i \) for \( i = 1, 2, \cdots, n \), which aligns with our expectations. We've illustrated these findings in Figure \ref{pie}. However, it's important to note that this approach is a preliminary trial and lacks robust theories. Recognizing its potential as an interesting research avenue, we have included it in the Appendix for further exploration and discussion.



\bibliographystyle{IEEEtran}
\bibliography{IEEEabrv,ref}

\begin{IEEEbiography}[{\includegraphics[width=1in,height=1in,keepaspectratio]{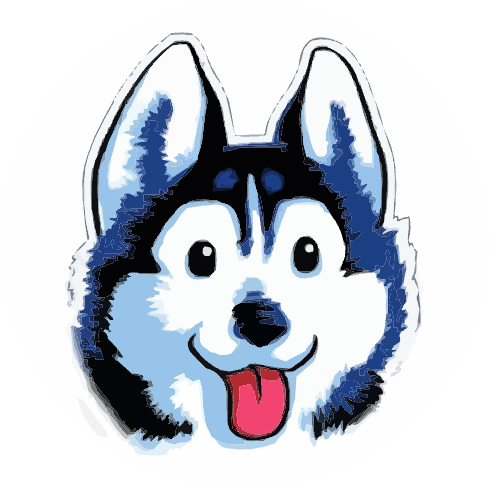}}]{Benhao Huang} is presently in his third year of undergraduate studies at Shanghai Jiao Tong University, majored in Computer Science. He holds a research internship position at the SJTU Interpretable ML Lab, where he works under the guidance of Prof. Quanshi Zhang. Recently, his research interests have been centered around Interpretable AI, Understanding Human Activity, Visual Reasoning, and Graph Learning. You can visit his website at huskydoge.github.io for more information. For any communication, he can be reached at hbh001098hbh@sjtu.edu.cn.
\end{IEEEbiography}




\end{document}